\documentclass[conference]{IEEEtran}

\IEEEoverridecommandlockouts
\usepackage{cite}
\usepackage{array}
\usepackage{multirow}
\usepackage{longtable}
\usepackage{rotating}
\usepackage{subfigure}
\usepackage{amsmath,amssymb,amsfonts}
\usepackage{algorithmic}
\usepackage{algorithm}
\usepackage{graphicx}
\usepackage{epstopdf}
\usepackage{textcomp}
\usepackage{xcolor}

\ifCLASSINFOpdf
\else
\fi
\begin{document}


\title{\Large \textbf{Boosting Federated Learning Convergence with Prototype Regularization}}


\author{
\IEEEauthorblockN{Yu Qiao\textsuperscript{1}, Huy Q. Le\textsuperscript{2} and Choong Seon Hong\textsuperscript{2}*}
\textsuperscript{1}
\textit{Department of Artificial Intelligence, Kyung Hee University, Yongin-si 17104, Republic of Korea}\\
\textsuperscript{2}
\textit{Department of Computer Science and Engineering, Kyung Hee University, Yongin-si 17104, Republic of Korea}\\
Email: \{qiaoyu, quanghuy69, cshong\}@khu.ac.kr
}


\twocolumn[\begin{@twocolumnfalse}       
\maketitle
\begin{abstract}{\normalfont\small\bfseries}         
As a distributed machine learning technique, federated learning (FL) requires clients to collaboratively train a shared model with an edge server without leaking their local data. However, the heterogeneous data distribution among clients often leads to a decrease in model performance. To tackle this issue, this paper introduces a prototype-based regularization strategy to address the heterogeneity in the data distribution. Specifically, the regularization process involves the server aggregating local prototypes from distributed clients to generate a global prototype, which is then sent back to the individual clients to guide their local training. The experimental results on MNIST and Fashion-MNIST show that our proposal achieves improvements of 3.3\% and 8.9\% in average test accuracy, respectively, compared to the most popular baseline FedAvg. Furthermore, our approach has a fast convergence rate in heterogeneous settings.

\vspace{0.5em}
\end{abstract}
\end{@twocolumnfalse}]
\IEEEpeerreviewmaketitle


\section{Introduction}
Over the past decade, significant progress has been made in deep learning, giving rise to a succession of impressive advancements in both large language models~\cite{zhang2023text,zhang2023complete,zhang2023one} and large vision models~\cite{kirillov2023segment,qiao2023robustness,zhang2023faster}. Concurrently, there has been an explosive growth of intelligent devices in distributed networks, generating a massive amount of raw data that requires processing. To address this situation, a powerful distributed machine learning paradigm called federated learning (FL) has been introduced.
FL is a powerful distributed machine learning paradigm that enables edge clients and servers to collaboratively train models without compromising the privacy of the underlying data \cite{mcmahan2017communication}. However, the data distribution of clients in federated scenarios is usually not Independent and Identically Distributed (non-IID), which can result in poor model performance \cite{qiao2023framework,qiao2023cdfed,qiao2023prototype,qiao2023knowledge}. There have been some studies exploring how to enhance the performance of FL in non-IID settings. FedAvg \cite{mcmahan2017communication} is the first optimization algorithm designed to address the challenge of data heterogeneity in federated scenarios. MPFed \cite{qiao2023framework} proposes a framework for addressing non-IID challenges in federated learning. The authors utilize a multi-prototype approach, where clients and the server adopt a typical FL approach for training, and model inference is based on the distance between local representations and the target prototype. CDFed \cite{qiao2023cdfed} proposes a dynamic federated learning paradigm based on Sharpley-value, where each client calculates its contribution to the global model in each iteration, thereby determining the probability of being selected for the next global iteration. Although effective, both of these strategies have high algorithmic complexity, which may pose difficulties in practical deployment. Furthermore, recent works in \cite{qiao2023prototype,qiao2023mp} propose a prototype-based model inference strategy, which is claimed to achieve faster convergence than the baseline FedAvg. However, they only adopted a prototype-based approach for model inference without modifying the federated learning process, which may not fully leverage the potential of prototypes.

In this paper, we propose to use prototype regularization to assist local training of clients, which makes better use of the prototype knowledge shared by each client. A prototype is a vector representation of a class space, and learning different prototype representations from clients can provide global information about the class space. Specifically, in each global iteration, the server aggregates the prototypes from all clients as the global prototype and transmits it along with the updated model parameters to the clients to assist them in their next local training. The main contributions can be summarized as follows:
\begin{itemize}
\item Our proposed federated training framework is based on prototypes, where both prototypes and model parameters are simultaneously optimized and updated in each iteration.
\item We propose a prototype-based regularization strategy, in which clients learn from the global prototype shared by all clients by minimizing the distance between their local representations and the global prototype.
\item Finally, based on experiments conducted on two widely-used benchmark datasets, MNIST\cite{lecun1998gradient} and Fashion-MNIST\cite{xiao2017fashion}, our proposed approach achieves significant improvements in test accuracy by 3.3\% and 8.9\%, respectively, compared to the most popular baseline method, FedAvg. Moreover, our approach exhibits a faster convergence rate in the heterogeneous setting.
\end{itemize}

\begin{figure}[t]
\centering
\includegraphics[width=0.45\textwidth]{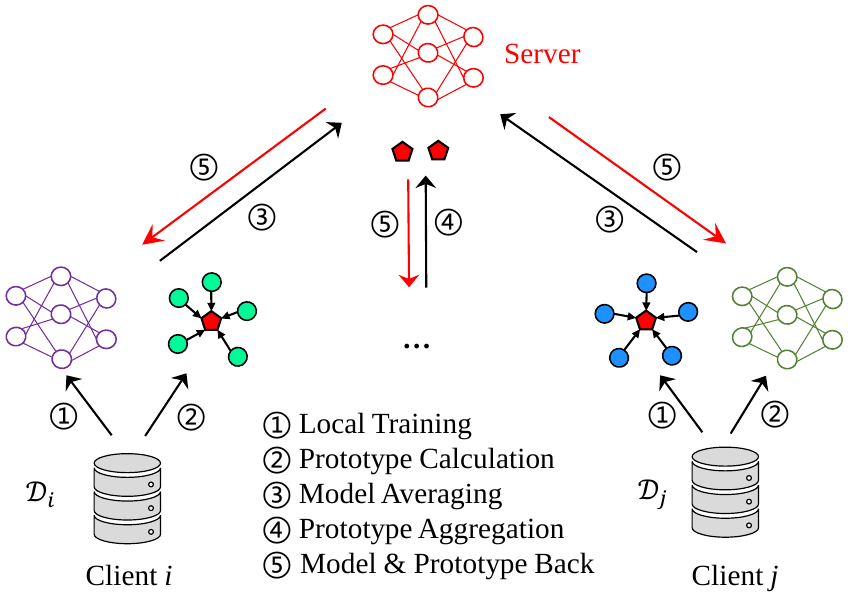}
\caption{The overview of the proposed FL framework that utilizes prototypes as a key component. In each global rounds, clients not only transmits their model parameters (step. 1 in this figure), but also their prototypes (step. 2) to the server for model averaging (step. 3) and prototype aggregation (step. 4). After the completion of each global round, the averaged model parameters and global prototypes are transmitted to all clients for the next round of training (step. 5).}
\label{System_model}
\end{figure}

\section{Proposed Framework}
\subsection{Problem Statement}
Consider a distributed set of clients $\phi$ with their private sensitive datasets $\mathcal{D}_i= {(\boldsymbol {x}_i, y_i)}$ of size $D_i$ in a distributed edge network. In a typical federated learning (FL) training process, clients and the edge server work together to train a shared model $\mathcal F(\omega; \boldsymbol {x}_i)$, where $\omega$ represents the model parameters of the global model, and $\boldsymbol {x}_i$ represents the feature vector of a specific client $i$. The goal is to minimize the loss function across clients with heterogeneous data, as proposed in \cite{qiao2023framework}:
\begin{equation} \label{global_loss}
    \mathop {\arg\min}_{\omega} {\mathcal{L}} (\omega) = \sum_{i \in \phi} \frac {D_i}{\sum_{i \in \phi}D_i} {\mathcal{L}_i} (\mathcal F(\omega; \boldsymbol {x}_i), y_i),
\end{equation}
where $y_i$ represents the label of a sample, and $\mathcal{L}_i$ denotes the cross-entropy loss of client $i$, respectively.

\subsection{Proposed Federated Learning Framework}
The typical federated training process involves distributing the global model to each client for local training, updating the model parameters independently, and aggregating the uploaded latest model parameters at the server before sending them back to the local sides for the next communication round. Our proposed framework follows a similar process, with the addition that in each global round, clients calculate their own prototypes for each class and send them to the server along with the model parameters for aggregation. Finally, the aggregated prototypes and the updated model parameters are sent back to local clients for further training. Figure \ref{System_model} provides an overview of our proposed framework.

\begin{algorithm}[t] 
    \caption{FedPR: Federated Prototype Regularization} 
    \label{alg:proto_reg} 
    \begin{algorithmic}[1]
        \REQUIRE ~~ \\
        Dataset $\mathcal{D}_i$, $\omega_i$
        \STATE \textbf{Model Training:}
        \STATE \textbf{Initialize $\omega^0$}, $\{\overline{y}\}$.
        \FOR{ $t$ = 1, 2, ..., $T$} 
            \FOR{ $i$ = 0, 1,..., $N$ \textbf{in parallel}}
                \STATE Local model updates using Eq. \eqref{obj:local_loss}.
                \STATE Local prototype calculation using Eq. \eqref{proto_define}.
                \STATE Global prototype aggregation using Eq. \eqref{proto_agg}.
            \ENDFOR           
        \ENDFOR
        \STATE \textbf{Model Inference:}
        \FOR{ each sample $i$ in testing dataset}
            \FOR{ each class $j$ in $\{\overline{y}\}$}
                \STATE Calculate the $\ell_2$ distance between $f_{e}(\omega_{e}; \boldsymbol {x}_i)$
                and $\overline{y}_j$
            \ENDFOR
            \STATE Make final predictions based on the smallest distance
        \ENDFOR
    \end{algorithmic}
\end{algorithm}

\subsection{Prototype-based Model Training}
In a typical deep learning model, there are two components: a feature extraction layer denoted by $f_e(\omega_e; \boldsymbol{x})$, and a decision-making layer denoted by $f_d(\omega_d; y)$. The feature extraction layer is designed to extract features from the input data, and the decision-making layer is responsible for utilizing these extracted features to make predictions or classifications.

\subsubsection{Prototype Calculation} According to the prototype definition proposed by \cite{qiao2023framework}, the prototype corresponding to the $j$-th class of client $i$-th can be calculated as follows:
\begin{equation} \label{proto_define}
\widetilde{y}_{i,j} = \frac{1}{D_{i,j}} \sum_{(x,y)\in \mathcal{D}_{i,j}}f_{e}(\omega_{e}; \boldsymbol {x}),
\end{equation}
where $\mathcal{D}_{i,j}$ refers to the distribution of samples in the $j$-th class belonging to client $i$-th, and $D_{i,j}$ denotes the size of this distribution. This formula aims to calculate the average of the feature representations obtained from the feature extraction layer $f_{e}(\omega_{e}; \boldsymbol {x})$ of all the samples in $\mathcal{D}_{i,j}$.

\subsubsection{Global Prototype Aggregation} The prototypes computed using the Eq. \ref{proto_define} can be aggregated by sending them to the server, and the aggregation process can be defined as follows \cite{qiao2023prototype}:
\begin{equation} \label{proto_agg}
\overline{y}_j = \frac{1}{N} \sum_{i\in N} \widetilde{y}_{i,j},
\end{equation}
where $N$ denotes the total number of clients in the network, and $\widetilde{y}_{i,j}$ represents the prototype vector for the $j$-th class of client $i$-th, as computed by Eq. \ref{proto_define}. The aggregation process involves computing the average of these prototypes from all clients that have samples of the $j$-th class, resulting in a global prototype vector $\overline{y}_j$. 

\subsubsection{Local Objective} The global representation for each class obtained through Eq. \ref{proto_agg} can be utilized for regularization purposes. Specifically, the loss function for each client can be defined as follows:
\begin{equation} \label{obj:local_loss}
\mathcal{L}_i(\omega_i) = \mathcal{L}_i (\mathcal F(\omega; \boldsymbol {x}_i), y_i) + \ell_2(f_{e}(\omega_{e};\boldsymbol {x}_i) - \overline{y}_j),
\end{equation}
where $\overline{y}_j$ belongs to $\mathbb Y$ which represents the set of global prototypes, $\mathcal{L}_i(\mathcal F(\omega; \boldsymbol {x}_i), y_i)$ is the supervised learning loss, and $\ell_2(f_{e}(\omega_{e};\boldsymbol {x}_i) - \overline{y}_j)$ calculates the $\ell_2$ distance between the local representation and the corresponding global prototype in $\mathbb Y$. Detailed training processes are presented in Algorithm \ref{alg:proto_reg}.

\section{Experiments}
\subsection{Dataset and Local Model} 
To compare our strategy with the baseline FedAvg, we use two benchmark datasets: MNIST \cite{lecun1998gradient} and Fashion-MNIST \cite{xiao2017fashion}. MNIST is a dataset composed of handwritten digits used for recognition tasks, while Fashion-MNIST consists of various clothing images. Both datasets share the same image size of 28x28x1 pixels. For both the MNIST and Fashion-MNIST datasets in our experiments, we employ a 4-layer CNN network comprising of 2 convolutional layers and 2 fully connected layers. Similar networks have also been used in previous research \cite{qiao2023framework,qiao2023cdfed}.

\begin{figure}[t]
\centering
\subfigure{\includegraphics[scale=0.45]{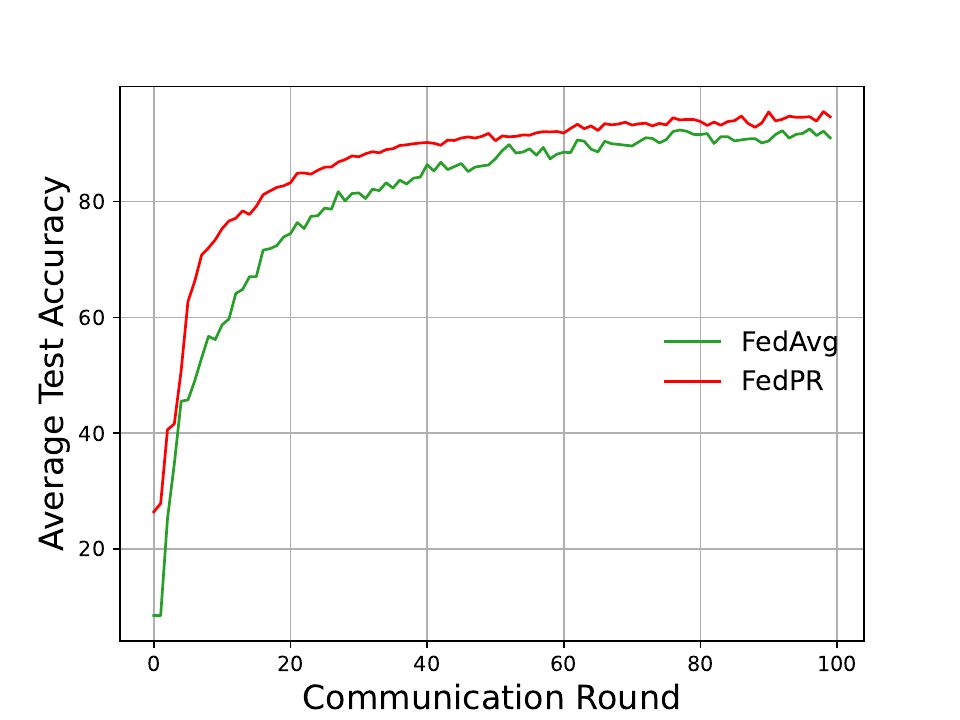}}
\caption{The top-1 average test accuracy of FedAvg and FedPR, on MNIST for different communication rounds, with the degree of data skewness set to $\alpha$ = 0.05.}
\label{mnist0.05}
\end{figure}

\begin{figure}[t]
\centering
\subfigure{\includegraphics[scale=0.45]{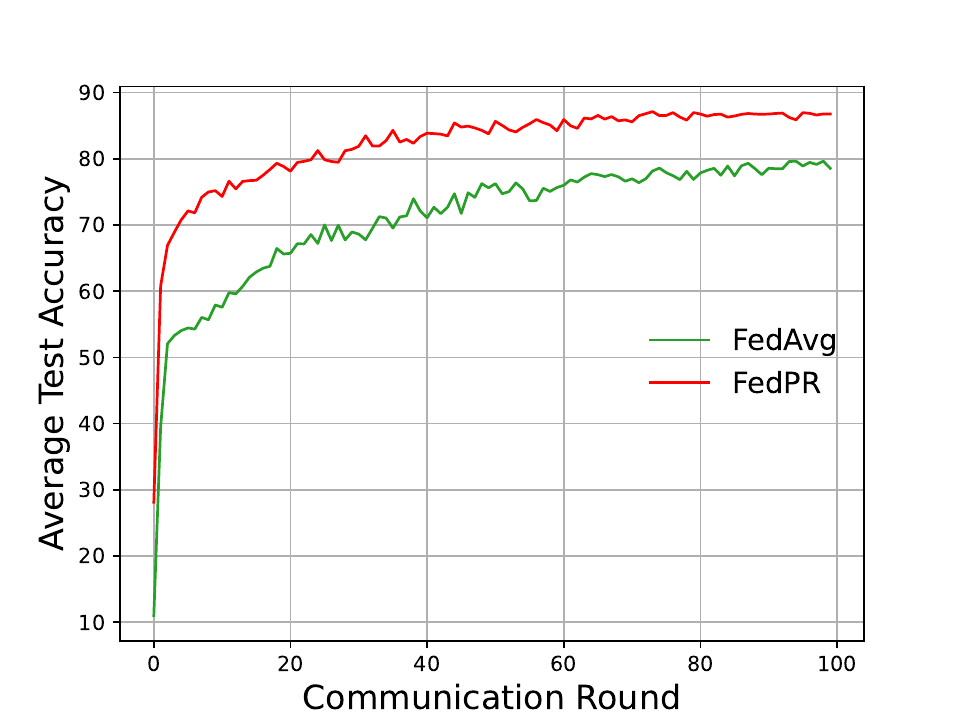}}
\caption{The top-1 average test accuracy of FedAvg and FedPR, on Fashion-MNIST for different communication rounds, with the degree of data skewness set to $\alpha$ = 0.05.}
\label{fashionmnist0.05}
\end{figure}

\subsection{Implementation Details}
For all experiments, we use 10 clients. We use the SGD optimizer for all baselines, and the SGD momentum is set to 0.5. The other training parameters for both MNIST and Fashion-MNIST datasets are set to $B$ = 8 and $\eta$ = 0.01, which represent the local batch size and learning rate, respectively. We denote local epoch as $E$, and set $E$ = 1 for MNIST and $E$ = 5 for Fashion-MNIST. To simulate the non-IID situation, we sample 2000 samples from the training dataset and distribute to all clients based on Dirichlet distribution Dir ($\alpha$) \cite{yurochkin2019bayesian}, where the smaller the value of $\alpha$, the more unbalanced the distribution of data is among clients.

\subsection{Accuracy and Communication Efficiency Comparison}
We compare our proposal with the most popular baseline \texttt{FedAvg} with Dir(0.05). The accuracy and communication efficiency for comparison of all methods on MNIST and Fashion-MNIST are shown in Fig.\ref{mnist0.05} and Fig.\ref{fashionmnist0.05}, respectively. It shows that our prototype-based regularization strategy achieves the higher test accuracy and faster convergence rate in each global communication round than the baseline FedAvg on both these two datasets. We take the test results of the last 10 rounds for both MNIST and Fashion-MNIST, and calculate the average test accuracy for our FedPR and FedAvg. For MNIST, the average accuracy of FedPR and FedAvg is 94.62\% and 91.57\% respectively, while for Fashion-MNIST, the average accuracy of FedPR and FedAvg is 86.05\% and 79.04\% respectively. In other words, our proposed approach achieves 3.3\% higher accuracy than \texttt{FedAvg} on MNIST and 8.9\% higher accuracy on Fashion-MNIST when $\alpha$ is set to 0.05.

\section{Conclusion}
In this article, we have proposed a FL framework based on prototype regularization to improve the model convergence rate. Specifically, we first introduce the prototype computation method, followed by the prototype aggregation method. Finally, we propose a prototype-based regularization strategy to regularize the local training of each client. Experimental results show that compared to the baseline FedAvg, our strategy can improve accuracy by 3.3\% and 8.9\% on MNIST and Fashion-MNIST, respectively, and achieve faster convergence speed. For future work, our proposal will be combined with other state-of-the-art methods and tested on more datasets.

\bibliographystyle{IEEEtran}
\bibliography{bib_global}

\end{document}